# Profiling Obese Subgroups in National Health and Nutritional Status Survey Data using Categorical Principal Component Analysis – A Case Study from Brunei Darussalam

Usman Khalil[a,*], Owais Ahmed Malik[a,b], Daphne Teck Ching Lai[a], Ong Sok King[c]

[a]School of Digital Science, Universiti Brunei Darussalam, Jalan Tungku Link, Gadong BE1410, Brunei Darussalam
[b]Institute of Applied Data Analytics, Universiti Brunei Darussalam, Jalan Tungku Link, Gadong BE1410, Brunei Darussalam
[c]Public Health Services, Ministry of Health, Brunei Darussalam and PAPRSB Institute of Health Sciences, Universiti Brunei Darussalam, Jalan Tungku Link, Gadong BE1410, Brunei Darussalam

| Article Info | ABSTRACT |
| --- | --- |
| *Keywords:*<br>NHANSS,<br>Machine Learning,<br>Categorical Principal Component Analysis, CATPCA,<br>NCD, Obesity. | National Health and Nutritional Status Survey (NHANSS) is conducted annually by the Ministry of Health in Negara Brunei Darussalam to assess the population's health and nutritional patterns and characteristics. The main aim of this study was to discover meaningful patterns (groups) from the obese sample of NHANSS data by applying the data reduction and interpretation techniques. The mixed nature of the variables (qualitative and quantitative) in the data set added the novelty to the study and accordingly, the Categorical Principal Component (CATPCA) technique was chosen to interpret the meaningful results. The relationships between obesity and the lifestyle factors like demography, socio-economic status, physical activity, dietary behavior, history of blood pressure, diabetes, etc., were determined based on the principal components generated by CATPCA. The results were validated with the help of the split method technique to counter-verify the authenticity of the generated groups. Based on the analysis and results, two subgroups were found in the data set, and the salient features of these subgroups have been reported. These results can be proposed for the betterment of the health care industry. |

## 1 Introduction

Obesity is one of the non-communicable diseases that is a condition of being overweight or a major nutritional disorder that has become a worldwide epidemic. Its growth has been projected at 40% in the upcoming decade [1]. It is often defined simply as a condition of abnormal or excessive fat accumulation in adipose tissue to the extent that health may be impaired [2]. Not only is being obese problematic, with that comes the risk and complications of other non-communicable diseases (NCDs) that can be life-threatening if not taken care of at the right time, e.g., hypertension/high blood pressure, diabetes, etc. [3]. Diseases can be communicable and non-communicable diseases; the names refer to the type as the diseases that are spread (by different means) from one individual to another such as pneumonia, malaria, hepatitis-A and C, HIV/AIDS, measles, etc. At the same time, the latter is not transmissible directly from one person to another e.g., obesity, cancer, heart disease, diabetes mellitus, cerebrovascular disease, hypertension, high blood pressure, high cholesterol levels, etc., [1, 2].

* Corresponding author. School of Digital Science, Universiti Brunei Darussalam, Jalan Tungku Link, Gadong BE1410, Brunei Darussalam.

*E-mail address:* uskhalil@gmail.com (Usman Khalil).

## 1.1 Obesity Prevalence in Brunei Darussalam

Brunei Darussalam is an oil & gas producing country and is one of the member countries in the ASEAN (Association of Southeast Asian Nations) organization. It is situated in South East Asia at the northern coast of Borneo Island, neighboring its borders directly with Malaysia. Its population is estimated at 417,200, with gross domestic products (GDP) per capita of USD 28,986 [5]. There has been a noticeable rise in non-communicable diseases (NCDs) as aforementioned [1], while obesity being one of NCDs, has been of major concern for the rise in its occurrence. The government has been targeting the management and prevention from the grassroots level to overcome this problem, including childhood obesity. It requires long-term strategies, and treating childhood obesity may likely help manage obese adults in the future [2, 4]. The prevalence of obesity and lifestyle factors affecting it has been the focus of the study, and the past studies have shown that it has been one of the major risk factors causing other non-communicable diseases such as diabetes and cardiovascular problems, etc. [6, 7, 8]. The threat of obesity-related NCD, especially chronic kidney diseases (CKD), are preventable by educating the population about the risks of being obese and prevention through a healthy lifestyle [4]. In 2014, an estimated over 600 million adults aged 18 years and above were obese worldwide [4].

## 1.2 ASEAN Strategic Framework on NCDs

Taking a step ahead, the Brunei Darussalam government has taken significant measures to handle the NCD-related issues in its population [3]. Following the World Health Organization (W.H.O) Global Action Plan to control the prevalence of non-communicable diseases (NCDs) and the ASEAN Strategic Framework on Health Development [9], the government has well anticipated the execution of the plan and have initiated a Multisectoral Action Plan on NCD (BruMap-NCD) 2013-2018 in order to control NCDs and related risk factors, it includes a ban on all kinds of smoking products in the country with 30% reduction in smoking prevalence and a 10% reduction in physical inactivity prevalence by 2018 from 2013 level [3]. According to this National Action Plan on the Prevention and Control of Noncommunicable Diseases (BruMAP-NCD) 2013-2018, Brunei's 1st National Nutritional Status Survey (NNSS) was carried out in the year 1997, around 32% percent of the population was overweight and 12% being obese among 20 years old and above [2, 5]. This obese percentage, in particular, was increased more than double to 27.2% in the year 2011 [3, 10, 11]. The current statistics show that around 61% of Bruneians are overweight and obese, which is the highest rate in ASEAN [2, 4, 5, 11]

## 1.3 National Health and Nutritional Status Survey Data

The data was provided by the ministry of health Brunei Darussalam which runs parallel with the Brunei Darussalam Household Expenditure Survey (HES) 2010/2011 implemented by the Department of Economic Planning and Development, Prime Minister's Office [10]. NHANSS acronym for National Health and Nutritional Status Survey is conducted annually to access the population's health and nutritional patterns and characteristics [10]. The data includes all the lifestyle aspects regarding demographics, socio-economic status, physical activity, and laboratory examinations. The Ministry of Health (MOH) designed the data collection process itself and was carried out in three phases. Sampling procedure, questionnaire development, database development, and testing.

Like others, NHANSS is also a cross-sectional survey aimed at the population aged from 5-to 75 years old with an initial target of 2184 participants from all the districts in Brunei Darussalam. All the health offices under the ministry of health were included for data collection, including Tertiary Care Hospitals, Health Offices in Districts, Health Clinics, and the Community Nutrition Centre were used as survey sites. Face-to-face interviews with parents and/or caregivers (for children) and participants themselves were conducted by trained dietitians/nutritionists and research assistants using a questionnaire booklet [10]. The measurements, including anthropometric indices such as weight, height, and waist circumference, were taken. Blood pressure readings were also noted for all respondents using standard methodology [13], while individuals aged 20 years and above were additionally asked for biochemical measurements. Before the final data collection, a test run was carried out in the survey procedures and questionnaire to have standardized data collection [3], [10], [13]

### 1.4 W.H.O Obesity Classification

W.H.O defined obesity as an accumulation of excessive body fats in tissues to the extent that health may be impaired. BMI measures it in $kg/m^2$ [2, 12]. It further defined the Overweight and obesity for adults as follows; BMI more than equal 25 $kg/m^2$ for Overweight; and BMI more than equal 30 $kg/m^2$ for Obese [2, 5]. Obesity has been further classified as BMI more than equal $30kg/m^2$ and includes an additional sub-division as BMI more than equal $30kg/m^2$ and less than equal 34.9 $kg/m^2$ for obese class-I, BMI more than equal $35kg/m^2$ and less than equal 39.9 $kg/m^2$ for obese class-II and BMI more than equal 40 $kg/m^2$ for obese class-III [2, 3, 8, 10, 12]. However, this classification does not completely consider the population-level heterogeneity and cannot identify the variations among obese individuals. There is evidence of the association of obesity with other factors, including demographics, nutrition habits, and individuals' physical activity [6, 7]. In our case, Body Mass Index (BMI) was calculated and inserted in the dataset as a variable feature to study the characteristics of obese people and the prevalence of obesity [1, 5]. This survey also used the same variable features like demographic status, diet patterns, physical activity together with the history of raised blood pressure, diabetes, and raised cholesterol with BMI measurements as done in the studies carried out in the past [7].

### 1.5 Categorical Principal Component

The field of study interested in developing computer algorithms to transform data into intelligent action is known as machine learning [16]. Machine learning techniques have been used to explore the details mentioned above, which have been of great importance to extract the useful knowledge from the data that normally is received from a group of individuals through a survey or a questionnaire or other health-related data collection techniques [15, 16]. Categorical Principal Component Analysis (CATPCA) is one of the techniques applied to the data sets with more variables to reduce the dimensions of the data set by making sure for as much variation as possible and, most importantly, applied to the set of qualitative and quantitative variables [18]. It has also helped the health care industry to classify the characteristics from patients for a particular disease and to use that information in order to improve the protocols and procedures for the better treatment of patients by the clinicians and, most importantly, for the betterment of humanity in general [7, 15, 17].

## 2 Methodology

The overview of the model methodology to carry out the study has been provided in Fig. 1. Data was taken from the NHANSS – 2017 provided by the Ministry of Health, Negara Brunei Darussalam, representing data collection and selecting the variables in the first step. The second step follows the data pre-processing for any missing values or normalization issues so that data can be applied with the machine learning techniques. The categorical principal component analysis (CATPCA) extracted the components by reducing the dimensions and classifying the data. In this research, the classification technique was tested, and the process of validation was carried out to check the authenticity of the generated results, while the last step concludes the interpretation by profiling the observed classes. The results, validation, and profiling steps were carried out to understand and present the intelligible data for reporting. Since the motive was to find the meaningful patterns, Fig. 1 shows the steps performed for identifying the subgroups of the obese in a given sample. The steps below mentioned were followed,

I. NHANSS ~ Obese Sample
II. Data Pre-Processing
III. Classification Method
IV. Interpretation & Analysis (Results & Discussion)
V. Results Validation

### 2.1 NHANSS ~ Obese Sample

Discussed in Section 1.3 and in order to study the characteristics of the obese population within the obese classes (I-II-III), the NHANSS data set (National Health and Nutritional Status Survey) was filtered with the number of people having BMI $\geq$ 30 kg/m$^2$. Out of the total sample of 2184 records, 449 were filtered with 20.55% percent, and the required set of variables was chosen. A subset data set was chosen from the NHANSS data, whereas all the variables were included based on evidence-based research on obesity [6, 13, 18]. Since the obese sample had mixed variable types, the data type measurement for the variables was defined as quantitative and qualitative. It also added to the study's novelty as not many studies for the obesity affecting factors have been carried out in the past with mixed variables data types. The level of measurement for quantitative variables was set as numeric, while for qualitative variables, the level of measurement was set either nominal (for not ordered data) or ordinal (for ordered data). At Step 3 in Fig. 1, the machine learning technique was applied once the data was pre-processed. The CATPCA (categorical principal component analysis) was chosen for this study because of its ability to handle qualitative and quantitative data.

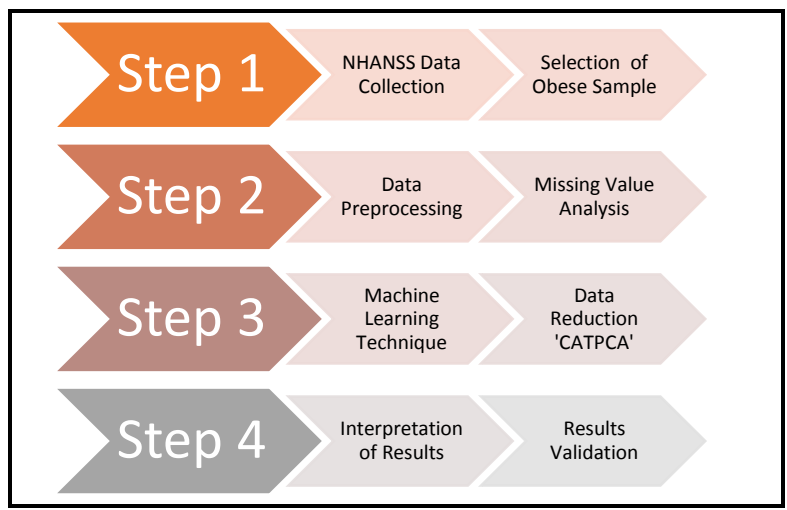

**Fig. 1.** Classification Model Methodology

2.2 Data Pre-processing

Like the other surveys, the NHANSS is a cross-sectional survey conducted among all age groups in all four districts. Fig. 2 lists the details for data collection [7], [10], [13]. As represented 67.70% of the data was collected from Brunei Muara, which is most densely populated, 17% from Kuala Belait, 12% from Tutong, and 3.30% from Temburong being lowest among all.

A comprehensive questionnaire was prepared to note down the critical information which was taken in several groups such as;

I. Demographics
II. Socio-economic status
III. Medical / Smoking Status
IV. Physical Activity Patterns
V. Anthropometric Measurements
VI. Multiple Dietary Patterns
VII. Bio-Chemical measurements on Adults and Children

The NHANSS data set with 2184 instances and 88 variables were pre-processed for missing values after missing value analysis. Since the data set was already imputed with the missing values, the data set was further analyzed. This sample was representative of obese individuals from all three classes of obesity. It had 449 instances with 88 variables (86 excluding BMI and Obesity factor as evaluating factors) for the CATPCA analysis. The level of measurement for all the variables was set to be ordinal while there were 14 numeric (Scale) variables whose normalization was taken care of by SPSS with a normal distribution. The data points were processed in SPSS Ver. 20 and since the obese NHANSS data set was used with 86 variables and 449 instances, the representation of number of variables (m) were $X_1, X_2, X_3......X_{86}$ i.e., m = 86 (e.g. $X_1$ = ageyears, $X_2$ = Urban, $X_3$ = DistCd and so on to………… $X_{86}$ = Salt93) while (n) represents the number of instances i.e., n=449 for obese sample.

Demographic variables were selected such as age, sex, marital status, etc., [20], for physical activity,

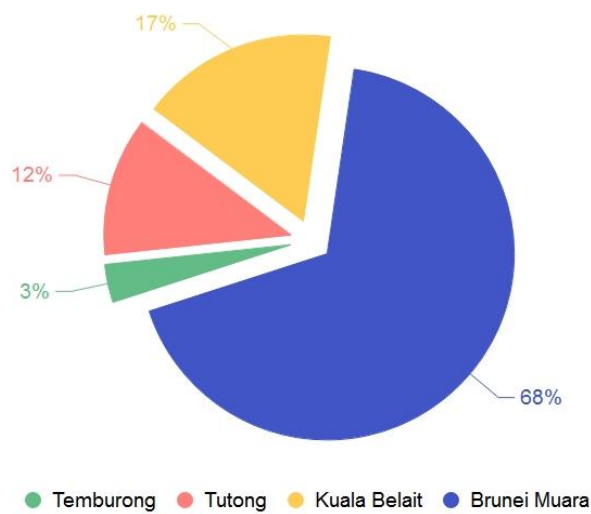

**Fig. 2**. Data Collection across the Districts of Brunei Darussalam

the recreational activities variables such as vigorous or moderate activities were selected, for sedentary characteristics, the time spent in watching TV and resting/reclining variables were taken as per their importance in the earlier studies [6, 19]. Age was reported as a continuous variable (quantitative), while sex, ethnicity, etc., were reported as categorical variables (qualitative) [15]. Time spent for vigorous/moderate activities, watching TV, and resting/reclining time were taken as continuous variables since they indicate the importance of sedentary characteristics with time spent on it [20]. Dietary intake was self-reported through a questionnaire provided to the subjects in NHANSS data collection [10]. It was reported in categorical variables format in levels from 1 to maximum 7, varying for different variables with numbers 666 for not known and 999 for not applicable, respectively [20].

2.3 Results Validation

For validation purposes, the split method was used, and the obese data set was divided into two data sets named train data set and test data set by a ratio of 70:30, which means 449 instances were divided by a ratio 314:135 respectively. First, the results were generated by applying CATPCA on train data set with 314 instances, and then these results were compared for validation of principal components by applying the same technique on the test data set (135 instances) later on. The descriptive statistics of obesity factors are presented in Table 1 with classes I, II, and III mentioned against obesity factors 1, 2, and 3 (1$^{st}$ column), respectively. The obesity factor was the representation of obesity classes in the data set. It can be seen that there were more cases from class I (187), with a percentage of 59.56% being the highest among the classes, followed by class II (85) with a percentage of 27.07% being second-highest and then class III (42) with a percentage of 13.37% being the lowest respectively.

2.4 Categorical Principal Component Analysis

Categorical principal component analysis (CATPCA) is applied to the data sets with more variables of mixed data types, i.e., qualitative and quantitative variables [21]. It reduces the dimensions of the data set by increasing variation as much as possible [18]. It is also referred to as Nonlinear Principal Component Analysis (PCA) [20], which works opposite of how PCA works. Nonlinear PCA reduces the observed variables to several uncorrelated variables [21]. If the measurement level of the variables is scaled to numeric, then PCA will be an alternative to CATPCA; therefore, it would not be wrong to say that CATPCA is an alternate analysis technique to PCA when the analysis required is to find the patterns of variations in a single data set of mixed data types [22]. When PCA handles mixed quantitative and qualitative data, the qualitative data must be quantified and is known as nonlinear PCA [18, 30, 31]. The CATPCA solution maximizes correlations of the object scores with each of the quantified variables for the number of components (dimensions) specified.

**Table 1.** Obese Sample ~ Train Data Set

| Obesity Factor | | Obesity Class | Frequency | Percent |
|---|---|---|---|---|
| Valid | 1[a] | Obese - Class I | 187 | 59.56% |
| | 2 | Obese - Class II | 85 | 27.07% |
| | 3 | Obese - Class III | 42 | 13.37% |
| | | Total | 314 | 100% |
| a. Mode | | | | |

The CATPCA application is only available in IBM® SPSS®. If applied to all variables that are declared multiple nominal, CATPCA produces an analysis equivalent to a multiple correspondence analysis (MCA) run on the same variables so CATPCA can be seen as a type of an MCA in which some of the variables are declared ordinal or nominal [22].

### 2.4.1 Component Extraction Methods

As discussed in the section above, one of the most important purposes of PCA / CATPCA methods is dimension reduction, and in order to achieve the purpose, some criterion has to be applied, whose method may follow the same principles to reduce the dimensions. Selecting only a few Principal Components (PCs) that share less of the variance may not help as this might result in selecting too few PCs and reducing the dimensions a lot. Similarly, selecting all the PCs will also be of no use just because they explain most of the variance of the data and may not help as this might result in selecting most or all the PCs and not reducing the dimensions at all. It may not fulfill the essence of the dimension reduction method.

### 2.4.2 Component Extraction Criterions

The principal components that share the maximum variance should be one benchmark to select and reduce the dimensions, but other defined criteria can be applied by looking at the data's nature. The different criteria available can be applied according to the nature of the data in the view. There are four types of criterion that can be used and are below mentioned;

I. EigenValue Criterion
II. The proportion of Variance Explained Criterion
III. Minimum Communality Criterion
IV. Scree Plot Criterion

#### 2.4.2.1 EigenValue Criterion

As per the eigenvalue criterion, a principal component must explain "one variable's worth, " which would mean that the PCs must have an eigenvalue of 1 at least. Eigenvalue Criterion may be best suited for the data sets with more than 20 and less than 50 variables if the data set has less than 20 variables. The criterion may choose too few principal components, and if the data set has more than 50 variables, then the criterion may choose too many principal components. In either case, it may not be feasible to analysis and later outline the characteristics of those chosen dimensions/components [18]. For instance, if there are the principal components PC1, PC2, and PC3 have eigenvalues $\lambda_1 = 1$, $\lambda_2 = 0.85, \lambda_3 = 0.075$ respectively then according to this criteria PC1 may be the only component retained, and the rest may be discarded. PC2 can also be retained as the eigenvalue is close to the threshold eigenvalue of 1, so in this case, two principal components may be retained, i.e., PC1 and PC2.

#### 2.4.2.2 The proportion of Variance Explained Criterion

This criterion mostly depends on the analyst who specifies the total number of principal components considering the total variability. The PCs must be selected one by one until the desired proportion of the variability explained is attained. The total proportion of the variability can be explained by Equation (2.1) below,

$$Z = \frac{\lambda_i}{m} \qquad (2.1)$$

Where;
- Z depicts the total variability in it,
- $\lambda_i$ depicts the eigenvalue for *ith* principal component
- m depicts the number of principal components

The equation represents the proportion of variability in Z, which is explained by the ratio of *ith* eigenvalue for the *ith* principal component to the number of variables. For instance, if a data set has ten variables applied with CATPCA results with eigenvalues against respective principal components and the first principal component has an eigenvalue of $\lambda_1 = 4.901$ then, as per equation 2.1, since there are ten variables (m), the first component may explain 4.901/10 = 49.01% of the shared variance among the predictor variables. Now, if the required percent of shared variance among the predictor variables is 85%, then more principal components may be added so that the desired number of components should attain the desired percent explained the variability.

#### 2.4.2.3 Minimum Communality Criterion

PCA / CATPCA does not present all the variance from the variables but only a proportion of the variance shared by the predictor variables. Communality plays an important role in extracting the proportion of a particular variable. Communality shows how beneficial the variable is for contributing to the CATPCA in terms of sharing the percent of the variance. If the variable shares less percent of the variance, it contributes less and vice versa, showing how beneficial the variables are to CATPCA. If it is required to keep a certain set of variables in the analysis, then most of the components with their weights are to be extracted so that the communality for each variable exceeds the minimum threshold of communality significance, i.e., 50%. It can be calculated as the sum of squared component weights for a given variable [16, 30].

#### 2.4.2.4 Scree Plot Criterion

The scree plot criterion has been used to extract the maximum number of components to work with. A Scree plot is a graphical representation of the eigenvalues against the component number and is very helpful in finding several components for further analysis. It always starts with a high value along the y-axis as it represents the eigenvalue for the first principal component explaining much of the shared variance, and later on, the line starts to dip along the x-axis as the eigenvalues for the rest of the principal components shares a lesser and lesser percentage of the variance. The significant knee of the line in two dimensions shows the number of principal components to be selected [22].

## 3  Results and Discussion

CATPCA was applied to the obese sample, which started with 0 iterations and the accounted variance of 87.045800 for all the variables at 0 iterations. Table 2 represents the iteration history of the CATPCA process. As depicted, the iterations stopped with an accounted variance of 87.108862 for all the variables at 100 iterations.

### 3.1  Principal Component Selection Criteria

As depicted in Table 2, the CATPCA algorithm finished iteration, and the eigenvalues were generated for all the 86 principal components with accounted variance shared by each of them. As noted in Table 3, the dimensions that share the maximum percent of the variance were to be selected. The first dimension shown in the table had an eigenvalue of 8.372, and it shares 9.74% of the total variance, which happens to be the highest percentage of shared variance among all the PCs, the eigenvalue was not very high, and that's because of the greater number of dimensions.

      Following the first dimension, the second-dimension shares 3.756% of the total variance similarly fifth until tenth and eleventh until thirty-first shares almost the same percentage of the total variance, i.e., $\geq 2$ and $\geq 1$ respectively. Since the percent was getting lower than 1%, the choice to choose dimensions was obvious, i.e., 31 dimensions. The next seven eigenvalues for the principal components were not very far from the threshold value of 1, so these components were also included. These dimensions shared 80.23% ≈ 80% percent of the total variance, which was not as good as required, but this was the maximum number of dimensions best suited for this data set. The relevant criterion to extract the principal components was checked and finalized in the next section.

**Table 2.** Iteration History ~ Train Data Set for Obese Sample

| Iteration Number | Variance Accounted For | | Loss | | |
|---|---|---|---|---|---|
| | Total | Increase | Total | Centroid Coordinates | Restriction of Centroid to Vector Coordinates |
| 0[a] | 87.045800 | .000004 | 7308.954200 | 7242.085225 | 66.868975 |
| 1 | 87.046360 | .000560 | 7308.953640 | 7242.085225 | 66.868415 |
| 2 | 87.047111 | .000750 | 7308.952889 | 7242.084079 | 66.868810 |
| . . . | | | Rows truncated | | |
| 98 | 87.108259 | .000308 | 7308.891741 | 7241.926309 | 66.965432 |
| 99 | 87.108563 | .000304 | 7308.891437 | 7241.925581 | 66.965856 |
| 100[b] | 87.108862 | .000300 | 7308.891138 | 7241.924856 | 66.966282 |

a. Iteration 0 displays the statistics of the solution with all variables, except variables with optimal scaling level Multiple Nominal, treated as numerical.
b. The iteration process stopped because the maximum number of iterations was reached.

**Table 3.** Model Summary ~ Train Data Set for Obese Sample

| Dim | Cronbach's Alpha | Variance Accounted For | | | | | |
|---|---|---|---|---|---|---|---|
| | | Total (Eigenvalue) | % of variance | Cum % | Eigenval Criterion | Proportion Variance Explained | Scree Plot |
| 1 | .891 | 8.372 | 9.74% | 9.74% | **9.74%** | **9.74%** | **9.74%** |
| 2 | .742 | 3.756 | 4.37% | 14.10% | **14.10%** | **14.10%** | **14.10%** |
| 3 | .702 | 3.261 | 3.79% | 17.89% | **17.89%** | **17.89%** | |
| 4 | .678 | 3.034 | 3.53% | 21.42% | **21.42%** | **21.42%** | |
| 5 | .659 | 2.866 | 3.33% | 24.75% | **24.75%** | **24.75%** | |
| 6 | .627 | 2.630 | 3.06% | 27.81% | **27.81%** | **27.81%** | |
| 7 | .608 | 2.505 | 2.91% | 30.73% | **30.73%** | **30.73%** | |
| 8 | .596 | 2.431 | 2.83% | 33.55% | **33.55%** | **33.55%** | |
| 9 | .548 | 2.181 | 2.54% | 36.09% | **36.09%** | **36.09%** | |
| 10 | .517 | 2.045 | 2.38% | 38.47% | **38.47%** | **38.47%** | |
| 11 | .504 | 1.992 | 2.32% | 40.78% | **40.78%** | **40.78%** | |
| 12 | .482 | 1.908 | 2.22% | 43.00% | **43.00%** | **43.00%** | |
| 13 | .476 | 1.890 | 2.20% | 45.20% | **45.20%** | **45.20%** | |
| 14 | .427 | 1.730 | 2.01% | 47.21% | **47.21%** | **47.21%** | |
| 15 | .384 | 1.612 | 1.87% | 49.08% | **49.08%** | **49.08%** | |
| 16 | .370 | 1.576 | 1.83% | 50.92% | **50.92%** | **50.92%** | |
| 17 | .336 | 1.497 | 1.74% | 52.66% | **52.66%** | **52.66%** | |
| 18 | .317 | 1.457 | 1.69% | 54.35% | **54.35%** | **54.35%** | |
| 19 | .310 | 1.442 | 1.68% | 56.03% | **56.03%** | **56.03%** | |
| 20 | .300 | 1.421 | 1.65% | 57.68% | **57.68%** | **57.68%** | |
| 21 | .274 | 1.372 | 1.59% | 59.28% | **59.28%** | **59.28%** | |
| 22 | .251 | 1.330 | 1.55% | 60.82% | **60.82%** | **60.82%** | |
| 23 | .240 | 1.311 | 1.52% | 62.35% | **62.35%** | **62.35%** | |
| 24 | .213 | 1.267 | 1.47% | 63.82% | **63.82%** | **63.82%** | |
| 25 | .197 | 1.242 | 1.44% | 65.26% | **65.26%** | **65.26%** | |
| 26 | .153 | 1.178 | 1.37% | 66.63% | **66.63%** | **66.63%** | |

| | | | | | | |
|---|---|---|---|---|---|---|
| 27 | .118 | 1.132 | 1.32% | 67.95% | **67.95%** | **67.95%** |
| 28 | .068 | 1.072 | 1.25% | 69.20% | **69.20%** | **69.20%** |
| 29 | .066 | 1.069 | 1.24% | 70.44% | **70.44%** | **70.44%** |
| 30 | .043 | 1.045 | 1.21% | 71.66% | **71.66%** | **71.66%** |
| 31 | .015 | 1.015 | 1.18% | 72.84% | **72.84%** | **72.84%** |
| 32 | -.040 | .962 | 1.12% | 73.95% | **73.95%** | **73.95%** |
| 33 | -.044 | .958 | 1.11% | 75.07% | **75.07%** | **75.07%** |
| 34 | -.070 | .935 | 1.09% | 76.16% | **76.16%** | **76.16%** |
| 35 | -.110 | .902 | 1.05% | 77.21% | **77.21%** | **77.21%** |
| 36 | -.138 | .880 | 1.02% | 78.23% | **78.23%** | **78.23%** |
| 37 | -.151 | .870 | 1.01% | 79.24% | **79.24%** | **79.24%** |
| 38 | -.180 | .849 | 0.99% | 80.23% | **80.23%** | **80.23%** |
| 39 | -.236 | .811 | 0.94% | 81.17% | | **81.17%** |
| 40 | -.239 | .809 | 0.94% | 82.11% | | **82.11%** |
| 41 | -.285 | .780 | 0.91% | 83.02% | | **83.02%** |
| 42 | -.355 | .740 | 0.86% | 83.88% | | **83.88%** |
| 43 | -.376 | .729 | 0.85% | 84.73% | | **84.73%** |
| 44 | -.408 | .713 | 0.83% | 85.56% | | **85.56%** |
| 45 | -.437 | .698 | 0.81% | 86.37% | | |
| 46 | -.462 | .687 | 0.80% | 87.17% | | |
| . . . | | | **Rows truncated** | | | |
| 85 | -226.810 | .004 | 0.00% | 101.29% | | |
| 86 | 0.000 | .000 | 0.00% | 101.29% | | |
| Total | 1.000ᵃ | 87.109 | | | | |
| a. Total Cronbach's Alpha is based on total Eigenvalue. | | | | | | |

### 3.2 Component Extraction Criterion

As discussed in Sections 2.4 to 2.4.2.4, the criterion were to be applied to the results to finalize the PCs so that profiling can be processed to know the characteristics of these PCs, respectively. The results presented by the algorithm in Table 3 give us the model summary for the percent of variance shared by all the PCs, and based on these eigenvalues; it was further evaluated for how many dimensions and how many PCs were to be extracted. Further presented in the table are the 86 dimensions for 100% of the shared variance in the data set. These dimensions were evaluated with criteria, and then the profiling of these PCs was finalized. The Eigenvalue criterion selected thirty-eight dimensions sharing an approximate 80.23% ≈ 80% of the total variance, which supports the theory of its tendency to extract more dimensions if variables in the dataset are > 50 variables. The proportion of variance explained criterion selected 44 dimensions sharing an approximate of 85.56% ≈ 86% of the total variance while the knee of the scree plot depicted in Fig. 3 suggested two principal components sharing an approximate of 14.10% ≈ 14% of the total variance to work with.

Since all the criteria were analyzed and applied to extract the exact number of components, finally, it was agreed to apply the eigenvalue criterion to extract the number of principal components. It was due to consideration of the said criterion for the variables that had eigenvalue ≥ 1 as it defines the one's variable worth; it also extracted a lesser number of principal components (38) comparatively with a reasonable percent of shared variance among the PCs, i.e., ≈ 80%. A total of 38 principal components reduced from a total of 86 principal components were considered, as shown in Table 3. Before profiling the principal components, the component weights also had to be evaluated.

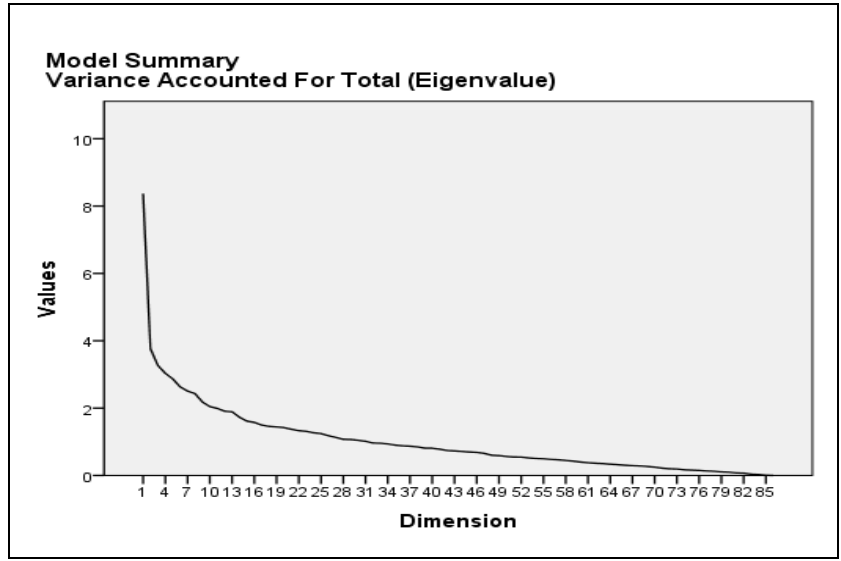

**Fig. 3.** Scree Plot Criterion for Component Selection ~ Obese Sample

Once the dimensions were chosen for the shared percent of variance by the principal components, it was time to evaluate and extract the components based on their factor weights. For evaluation of the component weights, the weight threshold value equal to +/-0.50 was considered to retain the component, which would define its contribution to CATPCA as a whole. The components that had component weight less than +/-0.50 were to be excluded as the decided threshold value in this research was +/-0.50 or values close it. Finally, 28 principal components in total were further excluded based on criterion and component weights, with ten principal components retained for profiling. The principal components extracted were PC6, 10-12, 14-26, and 28-38.

An overview of the extracted PCs concerning the factors within respective PCs has been presented in Fig. 4. As presented along the y-axis is the count of all factors within PC, while along the x-axis are extracted PCs. As discussed in Section 3.1, it is noticeable that PC1 has most of the variables count (PC1 = 16 variables), which correlates with the percent of variance shared among the PCs as normally the first PC has the maximum percent of the shared variance. As the PCs are added, the percent of the shared variance tends to decrease as seen in the figure below. PC1 has 16 variables count, PC2 has five variables count, PC3 has one variable count, and so on.

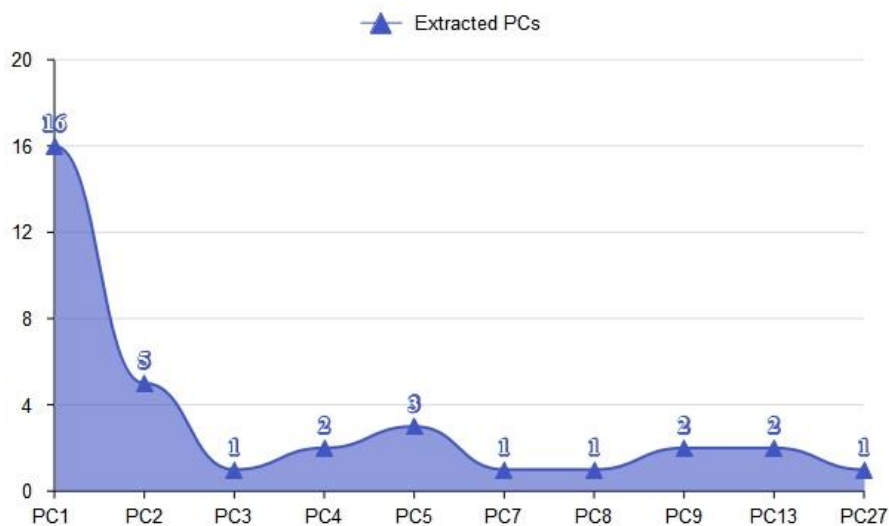

**Fig. 4.** CATPCA Final Extracted Components

### 3.2.1 Profiling the Principal Components

The most important part of the analysis is to profile the salient characteristics of each of the principal components so that useful knowledge can be extracted to align certain features. The obesity factor was kept as an evaluation factor to know the characteristics of obese individuals. Table 4 below presents all the factors allotted to respective PCs, and later on, all the principal components are discussed to note the salient features of all the PCs, respectively.

#### 3.2.1.1 Principal Component 1

PC1 presented in Table 4 is composed largely of the "block group size" variables, namely, *history of raised blood pressure:* tablets taken (53a), diet (53b), lose weight (53c), stop smoking (53d), start exercise (53e), *history of diabetes:* blood sugar measured in past 12 months (56), tablets taken (60b), diet (60c), weight loss (60d), smoking (60e), start exercise (60f), *history of raised blood cholesterol:* tablets taken (65a), diet (65b), weight loss (65c), stop smoking (65d), start exercise (65e), all have large values referred to an as high level. The values presented in Table 4 show that these variables were right-skewed. It means most of the individuals were not receiving any advice from the doctor or treatment in terms of tablets, prescribed diet plan, weight loss, stop smoking habit, to start or stop the exercise as far as the history of raised blood pressure was concerned and same trend was observed for a history of diabetes and a history of high blood cholesterol. PC1 shares the maximum percent of variance by factors, and the salient characteristics showed that this component belonged to healthy individuals with no history of raised blood pressure, diabetes, and high blood cholesterol.

**Table 4.** Components Extraction ~ CATPCA

| Sr.# | Variables | PC. No. | PC Wts | Sr. # | Variables | PC. No. | PC Wts |
|---|---|---|---|---|---|---|---|
| 1 | ageyears | PC3 | 0.511090985 | 18 | Diet60c | PC1 | 0.669637075 |
| 2 | DistCd | PC4 | -0.484673219 | 19 | Wtls60d | PC1 | 0.760748154 |
| 3 | Gndr6 | PC2 | -0.775430392 | 20 | Smk60e | PC1 | 0.637104831 |
| 4 | Relgn10 | PC7 | -0.519706542 | 21 | Exer60f | PC1 | 0.760748154 |
| 5 | Elec13 | PC13 | 0.515694454 | 22 | HadCh64 | PC5 | 0.600427088 |
| 6 | Pwtr14 | PC13 | 0.552189942 | 23 | Tab65a | PC1 | 0.634583711 |
| 7 | Cusmk24 | PC2 | -0.523144934 | 24 | Diet65b | PC1 | 0.641509808 |
| 8 | Smkls33 | PC4 | 0.52311049 | 25 | Lswt65c | PC1 | 0.790191235 |
| 9 | HadBP52 | PC5 | 0.531008267 | 26 | Smk65d | PC1 | 0.623978676 |
| 10 | Tab53a | PC1 | 0.587766619 | 27 | Anem68f | PC27 | -0.474800944 |
| 11 | Diet53b | PC1 | 0.516720969 | 28 | Imge69 | PC9 | 0.545241798 |
| 12 | Lswt53c | PC1 | 0.657802241 | 29 | WtNow70 | PC9 | 0.533603153 |
| 13 | Smk53d | PC1 | 0.569551547 | 30 | Wt73 | PC2 | 0.696848744 |
| 14 | Exer53e | PC1 | 0.657887243 | 31 | Ht74 | PC2 | 0.783204775 |
| 15 | BdSgr56 | PC1 | 0.537192999 | 32 | Wst75 | PC2 | 0.599904753 |
| 16 | HadDM57 | PC5 | 0.61081779 | 33 | NasiK90 | PC8 | -0.464404107 |
| 17 | Tab60b | PC1 | 0.57700017 | 34 | Exer65e | PC1 | 0.782187232 |

#### 3.2.1.2 Principal Component 2

Table 4 depicts PC2, which is about demographic status, socio-economic status, smoking status, recreational activity, and body image. It showed that most of the individuals were *female* (6). Mostly never smoked any tobacco products such *as cigarettes, cigars, or pipes in recent times* (24). The body image showed that the individuals in this principal component had *heavyweights* (73) and *heights* (74) along with their *waists* (75).

#### 3.2.1.3 Principal Component 3

PC3 presents the demographic characteristics of the sample as presented in Table 4. The age in years and values were noted high and increasing concerning obesity which means this variable had contributed well in CATPCA and that most of the individuals were *elderly aged* (1).

#### 3.2.1.4 Principal Component 4

PC4 presents the sample's demographic, socio-economic, smoking, and health characteristics as presented in Table 4. The level states that most individuals lived in the main *districts* (DistCd) of Brunei Darussalam. They were *Brunei Citizens* (8) and did not *smoke daily* (33) along with *tobacco products such as cigarettes, cigars, or pipes*.

#### 3.2.1.5 Principal Component 5

PC5 in Table 4 presents physical activity status, history of raised blood pressure, and blood cholesterol. It showed high values for all the factors. The individuals in this PC did not know about being told by a doctor or health worker for having *high bp or hypertension* (52); similarly, they have never been told by a doctor or health worker for *high blood sugar* level or *diabetes* (57) during the past 12 months. They also had never been told by a doctor or health worker about *high blood cholesterol* (64) during the past 12 months.

#### 3.2.1.6 Principal Component 7, 8, 9

PC7, 8, and 9 in Table 4 present the individuals' demographic status, body image, and short food frequency status. PC7 depicts that most individuals were Muslim belonging to the *religion* (10) Islam.

PC8 & PC9 in Table 4 presented the body image and short food frequency status of the obese sample. Most of the individuals considered themselves *Overweight* (69) and were not satisfied with their *body weights* (70), while most of the people were used to eating *nasi kato*k (90) and *Chicken Tail / Wings / Skin* (91) twice a week.

#### 3.2.1.7 Principal Component 13

PC13 in Table 4 depicts socio-economic status with high values. It depicts that most individuals had *electricity* and *water piped supply* (13 and 14) to their houses.

#### 3.2.1.8 Principal Component 27

PC27 in Table 4 represents the health status which showed that most of the individuals were suffering from *anemia* (68f) as far as health was concerned.

### 3.3 Minimum Communalities Criterion

As discussed in Section 2.4.2.3 and Table 3, the variable that shares less communality means shares less of its common variability among the variables, and contribution to the CATPCA is also considered lesser. At first, the finalized PCs were compiled with respect to the factor variable weights (≥ +/- 0.50), as highlighted in Table 5. The communality values showed the contributing factor variables. 35-factor variables out of the total 86 variables met the criteria, and the rest were omitted. At the second step, all the respective PCs' weights (Table 5) according to the communality criterion were calculated with their squared weights.

Table 6 depicts the squared weights for all the factor variables in respective PCs. It shows the squared component loadings for the 25-factor variables that met the criteria by having a communality significance value ≥ 50% showing their contribution to CATPCA. It means these are the final set of factor variables that have contributed well to the algorithm as a whole. CATPCA classified the NHANSS data in two subgroups; one subgroup was presented with left-skewed distribution while the other was presented with right-skewed distribution, which means that the most prevalent conditions with respect to obesity by variables were either detected or undetected. The variables that presented the communality significance more than the threshold (50%) value were the variables that helped gain knowledge about the salient characteristics from the NHANSS obese sample. As discussed above, the generic details of these variables are below mentioned for reference.

- Demographic and Socio-Economic Characteristics
- Smoking Characteristics
- History of Raised Blood Pressure
- History of Diabetes Mellitus
- History of High Blood Cholesterol and,
- Anthropometric Characteristics

**Table 5.** Components Loadings ~ Train Data Set for Obese Sample

| Variables | Dimension | | | | | | | | | |
|---|---|---|---|---|---|---|---|---|---|---|
| | 1 | 2 | 3 | 4 | 5 | 7 | 8 | 9 | 13 | 27 |
| ageyears | -.382 | -.158 | **.511** | -.079 | -.168 | .087 | .127 | -.032 | -.053 | .045 |
| DistCd | .047 | .057 | .221 | **-.485** | .394 | -.335 | -.008 | .038 | .052 | .133 |
| Gndr6 | .222 | **-.775** | .016 | -.124 | -.040 | -.108 | .143 | -.039 | -.013 | -.003 |
| Relgn10 | .056 | .201 | .154 | -.153 | .106 | **-.520** | -.128 | .189 | -.221 | .085 |
| Elec13 | .054 | -.006 | .072 | -.118 | -.044 | -.316 | .139 | .196 | **.516** | .025 |
| Pwtr14 | .030 | .030 | .026 | -.005 | -.022 | -.220 | .187 | .132 | **.552** | -.019 |
| Cusmk24 | .075 | **-.523** | .116 | .110 | .205 | -.121 | .003 | -.040 | -.060 | .020 |
| Smkls33 | -.024 | -.082 | -.087 | **.523** | .362 | .054 | .485 | .250 | -.042 | .023 |
| HadBP52 | .038 | -.013 | -.033 | -.375 | **.531** | .341 | -.004 | .072 | .015 | -.069 |
| Tab53a | **.588** | .140 | -.395 | -.101 | -.131 | .013 | -.026 | .061 | .087 | .017 |
| Diet53b | **.517** | .198 | -.377 | -.147 | -.176 | -.005 | -.011 | .150 | .049 | -.020 |
| Lswt53c | **.658** | .045 | -.342 | -.094 | -.053 | -.149 | .032 | -.073 | .125 | .030 |
| Smk53d | **.570** | -.198 | .011 | .193 | .165 | -.056 | .000 | -.082 | .020 | -.084 |
| Exer53e | **.658** | .061 | -.383 | -.100 | -.066 | -.122 | -.001 | -.039 | .145 | .017 |
| BdSgr56 | **.537** | .185 | -.140 | -.030 | .074 | .007 | .019 | .116 | .048 | .003 |
| HadDM57 | .092 | .041 | .012 | -.443 | **.611** | .327 | .000 | .102 | .042 | -.001 |
| Tab60b | **.577** | .232 | .374 | .074 | -.110 | .307 | .063 | .039 | .119 | .018 |
| Diet60c | **.670** | .197 | .370 | .061 | -.108 | .249 | .048 | .071 | .059 | .005 |
| Wtls60d | **.761** | .121 | .383 | .126 | -.015 | .125 | -.007 | -.024 | .113 | .026 |
| Smk60e | **.637** | -.121 | .361 | .150 | .172 | -.046 | -.141 | -.052 | -.008 | -.043 |
| Exer60f | **.761** | .121 | .383 | .126 | -.015 | .125 | -.007 | -.024 | .113 | .026 |
| HadCh64 | .087 | .058 | .010 | -.416 | **.600** | .341 | .012 | .080 | .046 | -.032 |
| Tab65a | **.635** | .158 | .080 | -.007 | -.137 | .077 | .131 | -.023 | -.150 | -.050 |
| Diet65b | **.642** | .165 | .016 | -.066 | -.150 | -.034 | .194 | .042 | -.193 | -.013 |
| Lswt65c | **.790** | .027 | .084 | -.001 | -.042 | -.124 | .125 | -.103 | -.164 | .001 |
| Smk65d | **.624** | -.178 | .251 | .126 | .170 | -.133 | -.093 | -.076 | -.113 | -.075 |
| Anem68f | -.031 | .137 | -.123 | .053 | -.008 | .070 | -.151 | .198 | .058 | **-.475** |
| Imge69 | .058 | -.077 | -.122 | .105 | -.047 | -.092 | .064 | **.545** | -.143 | .021 |
| WtNow70 | .075 | -.086 | -.164 | .112 | -.046 | -.096 | .067 | **.534** | -.130 | .029 |
| Wt73 | -.340 | **.697** | .082 | .140 | .035 | -.049 | -.019 | .033 | -.134 | -.041 |
| Ht74 | -.234 | **.783** | .125 | .147 | .111 | .014 | -.156 | .032 | -.033 | -.029 |
| Wst75 | -.439 | **.600** | .063 | .066 | -.023 | .008 | .026 | .077 | -.167 | -.056 |
| NasiK90 | .006 | -.147 | .210 | .101 | .101 | .039 | **-.464** | .212 | -.152 | -.156 |
| Exer65e | **.782** | .043 | .069 | -.025 | -.041 | -.142 | .142 | -.108 | -.181 | -.017 |

**Table 6.** Squared Components Loadings Communality ~ Train Data Set for Obese Sample

| Variables | Dimension | | | | | | | | | | Comm. Criteria |
|---|---|---|---|---|---|---|---|---|---|---|---|
| | 1 | 2 | 3 | 4 | 5 | 7 | 8 | 9 | 13 | 27 | Sqrd.Sum |
| **ageyears** | **.146** | **.025** | **.261** | **.006** | **.028** | **.008** | **.016** | **.001** | **.003** | **.002** | **.497** |
| **DistCd** | **.002** | **.003** | **.049** | **.235** | **.155** | **.112** | **.000** | **.001** | **.003** | **.018** | **.579** |
| **Gndr6** | **.049** | **.601** | **.000** | **.015** | **.002** | **.012** | **.021** | **.002** | **.000** | **.000** | **.702** |
| **Relgn10** | **.003** | **.040** | **.024** | **.023** | **.011** | **.270** | **.016** | **.036** | **.049** | **.007** | **.480** |
| Elec13 | .003 | .000 | .005 | .014 | .002 | .100 | .019 | .038 | .266 | .001 | .448 |
| Pwtr14 | .001 | .001 | .001 | .000 | .000 | .048 | .035 | .017 | .305 | .000 | .409 |
| Cusmk24 | .006 | .274 | .014 | .012 | .042 | .015 | .000 | .002 | .004 | .000 | .367 |
| **Smkls33** | **.001** | **.007** | **.008** | **.274** | **.131** | **.003** | **.235** | **.062** | **.002** | **.001** | **.722** |
| **HadBP52** | **.001** | **.000** | **.001** | **.141** | **.282** | **.116** | **.000** | **.005** | **.000** | **.005** | **.552** |
| **Tab53a** | **.345** | **.019** | **.156** | **.010** | **.017** | **.000** | **.001** | **.004** | **.007** | **.000** | **.561** |
| **Diet53b** | **.267** | **.039** | **.142** | **.021** | **.031** | **.000** | **.000** | **.022** | **.002** | **.000** | **.526** |
| **Lswt53c** | **.433** | **.002** | **.117** | **.009** | **.003** | **.022** | **.001** | **.005** | **.016** | **.001** | **.608** |
| Smk53d | .324 | .039 | .000 | .037 | .027 | .003 | .000 | .007 | .000 | .007 | .446 |
| **Exer53e** | **.433** | **.004** | **.147** | **.010** | **.004** | **.015** | **.000** | **.002** | **.021** | **.000** | **.635** |
| BdSgr56 | .289 | .034 | .020 | .001 | .005 | .000 | .000 | .013 | .002 | .000 | .365 |
| **HadDM57** | **.008** | **.002** | **.000** | **.196** | **.373** | **.107** | **.000** | **.010** | **.002** | **.000** | **.699** |
| **Tab60b** | **.333** | **.054** | **.140** | **.005** | **.012** | **.094** | **.004** | **.002** | **.014** | **.000** | **.659** |
| **Diet60c** | **.448** | **.039** | **.137** | **.004** | **.012** | **.062** | **.002** | **.005** | **.003** | **.000** | **.712** |
| **Wtls60d** | **.579** | **.015** | **.147** | **.016** | **.000** | **.016** | **.000** | **.001** | **.013** | **.001** | **.786** |
| **Smk60e** | **.406** | **.015** | **.130** | **.022** | **.030** | **.002** | **.020** | **.003** | **.000** | **.002** | **.630** |
| **Exer60f** | **.579** | **.015** | **.147** | **.016** | **.000** | **.016** | **.000** | **.001** | **.013** | **.001** | **.786** |
| **HadCh64** | **.008** | **.003** | **.000** | **.173** | **.361** | **.116** | **.000** | **.006** | **.002** | **.001** | **.671** |
| **Tab65a** | **.403** | **.025** | **.006** | **.000** | **.019** | **.006** | **.017** | **.001** | **.023** | **.003** | **.502** |
| **Diet65b** | **.412** | **.027** | **.000** | **.004** | **.022** | **.001** | **.037** | **.002** | **.037** | **.000** | **.544** |
| **Lswt65c** | **.624** | **.001** | **.007** | **.000** | **.002** | **.015** | **.016** | **.011** | **.027** | **.000** | **.702** |
| **Smk65d** | **.389** | **.032** | **.063** | **.016** | **.029** | **.018** | **.009** | **.006** | **.013** | **.006** | **.579** |
| **Exer65e** | **.612** | **.002** | **.005** | **.001** | **.002** | **.020** | **.020** | **.012** | **.033** | **.000** | **.706** |
| Anem68f | .001 | .019 | .015 | .003 | .000 | .005 | .023 | .039 | .003 | .225 | .334 |
| Imge69 | .003 | .006 | .015 | .011 | .002 | .009 | .004 | .297 | .020 | .000 | .368 |
| WtNow70 | .006 | .007 | .027 | .013 | .002 | .009 | .004 | .285 | .017 | .001 | .371 |
| **Wt73** | **.115** | **.486** | **.007** | **.020** | **.001** | **.002** | **.000** | **.001** | **.018** | **.002** | **.652** |
| **Ht74** | **.055** | **.613** | **.016** | **.022** | **.012** | **.000** | **.024** | **.001** | **.001** | **.001** | **.745** |
| **Wst75** | **.193** | **.360** | **.004** | **.004** | **.001** | **.000** | **.001** | **.006** | **.028** | **.003** | **.599** |
| NasiK90 | .000 | .022 | .044 | .010 | .010 | .002 | .216 | .045 | .023 | .024 | .396 |

### 3.4 Validation of Principal Components ~ Obese Sample

As discussed in Section 2.3, the test data set taken from 449 instances (obese sample) was divided with a ratio (70:30) of 314:135, respectively. CATPCA generated the results on the test data set, and then these results were compared for validation of principal components with those already generated aforementioned. It was noticeable that the results generated by the train data set did not show much difference with respect to the selection and extraction of components for further evaluation. The process started with 0 iterations and ended at 100 iterations. As shown in Table 7, the shared variance was noted at 87.999 as a whole by the CATPCA.

**Table 7.** Model Summary for Test Data Set ~ Obese Sample

| Dim | Cronbach's Alpha | Variance Accounted For | | | | | |
|---|---|---|---|---|---|---|---|
| | | Total (Eigenval) | % of variance | Cum % | Eigenvalue Criterion | Proportion Variance Explained | Scree Plot |
| 1 | .910 | 9.954 | 11.44% | 11.44% | **11.44%** | **11.44%** | **11.44%** |
| 2 | .783 | 4.421 | 5.08% | 16.52% | **16.52%** | **16.52%** | **16.52%** |
| 3 | .763 | 4.057 | 4.66% | 21.19% | **21.19%** | **21.19%** | |
| 4 | .702 | 3.262 | 3.75% | 24.94% | **24.94%** | **24.94%** | |
| 5 | .668 | 2.940 | 3.38% | 28.32% | **28.32%** | **28.32%** | |
| 6 | .657 | 2.851 | 3.28% | 31.59% | **31.59%** | **31.59%** | |
| 7 | .650 | 2.800 | 3.22% | 34.81% | **34.81%** | **34.81%** | |
| 8 | .641 | 2.727 | 3.13% | 37.95% | **37.95%** | **37.95%** | |
| 9 | .616 | 2.559 | 2.94% | 40.89% | **40.89%** | **40.89%** | |
| 10 | .577 | 2.327 | 2.68% | 43.56% | **43.56%** | **43.56%** | |
| 11 | .541 | 2.151 | 2.47% | 46.03% | **46.03%** | **46.03%** | |
| 12 | .521 | 2.061 | 2.37% | 48.40% | **48.40%** | **48.40%** | |
| 13 | .518 | 2.048 | 2.35% | 50.76% | **50.76%** | **50.76%** | |
| 14 | .493 | 1.951 | 2.24% | 53.00% | **53.00%** | **53.00%** | |
| 15 | .451 | 1.801 | 2.07% | 55.07% | **55.07%** | **55.07%** | |
| 16 | .439 | 1.765 | 2.03% | 57.10% | **57.10%** | **57.10%** | |
| 17 | .408 | 1.675 | 1.93% | 59.02% | **59.02%** | **59.02%** | |
| 18 | .387 | 1.618 | 1.86% | 60.88% | **60.88%** | **60.88%** | |
| 19 | .368 | 1.573 | 1.81% | 62.69% | **62.69%** | **62.69%** | |
| 20 | .334 | 1.493 | 1.72% | 64.41% | **64.41%** | **64.41%** | |
| 21 | .305 | 1.433 | 1.65% | 66.05% | **66.05%** | **66.05%** | |
| 22 | .268 | 1.360 | 1.56% | 67.62% | **67.62%** | **67.62%** | |
| 23 | .242 | 1.313 | 1.51% | 69.13% | **69.13%** | **69.13%** | |
| 24 | .225 | 1.286 | 1.48% | 70.61% | **70.61%** | **70.61%** | |
| 25 | .205 | 1.252 | 1.44% | 72.04% | **72.04%** | **72.04%** | |
| 26 | .174 | 1.209 | 1.39% | 73.43% | **73.43%** | **73.43%** | |
| 27 | .152 | 1.178 | 1.35% | 74.79% | **74.79%** | **74.79%** | |
| 28 | .090 | 1.099 | 1.26% | 76.05% | **76.05%** | **76.05%** | |
| 29 | .062 | 1.066 | 1.23% | 77.28% | **77.28%** | **77.28%** | |
| 30 | .046 | 1.049 | 1.21% | 78.48% | **78.48%** | **78.48%** | |
| 31 | .000 | 1.000 | 1.15% | 79.63% | **79.63%** | **79.63%** | |
| 32 | -.040 | .962 | 1.11% | 80.74% | **80.74%** | **80.74%** | |
| 33 | -.095 | .914 | 1.05% | 81.79% | **81.79%** | **81.79%** | |
| 34 | -.151 | .870 | 1.00% | 82.79% | **82.79%** | **82.79%** | |
| 35 | -.180 | .848 | 0.97% | 83.76% | | **83.76%** | |
| 36 | -.203 | .833 | 0.96% | 84.72% | | **84.72%** | |
| 37 | -.242 | .808 | 0.93% | 85.65% | | | |
| 38 | -.323 | .758 | 0.87% | 86.52% | | | |
| . . . | | | **Rows truncated** | | | | |
| 87 | -86.921 | .012 | 0.01% | 89.71% | | | |
| Total | 1.000[a] | 87.999 | | | | | |
| a. Total Cronbach's Alpha is based on total Eigenvalue. | | | | | | | |

The model summary was generated against the eigenvalues representing the percent of variance shared among the principal components. To evaluate these PCs and to know whether this test data set had also generated the same number of PCs, a comparison had to be made to indicate whether these results for the data set as a whole are generalized or not, so the results can be reported as Valid or Invalid. The eigenvalues starting from PC1, both the data sets, train, and test data sets, almost shared the same percentage of variance reported as 8.372 and 9.954, respectively. Similarly, for PC2, the eigenvalues were reported as 3.756 and 4.421, respectively. For PC3, the eigenvalues were reported as 3.261 and 4.057, respectively, and so on. Here, it is wise to compare the criterion results from train and test data sets to see if the reported results were the same as those of eigenvalues or if they differ by a great number. If there were a minimal difference in the number of selected PCs or shared variance, it would validate the results, but if vice versa, then the validation would be reported as invalid as far as the reporting the results were concerned. Since the eigenvalue criteria were finalized for the train data set, the results with respect to the eigenvalue criterion generated by the test data set were checked and compared for validation.

3.4.1 EigenValue Criterion ~ Test Data Set

The results presented in Table 7 showed the same trend of extracting more PCs in terms of dimensions as the data set had more than 50 variables, and hence the criterion suggests extracting exactly 31 dimensions that have eigenvalue ≥ 1 also the next three proceeding dimensions that had eigenvalues close to 1, i.e., ≤ 0.85, were added. A total of 34 dimensions were suggested by this criterion, sharing approximately 82.79% ≈ 83% of the total variance, which again supported the theory of its tendency to extract more dimensions (if variables in the data set are > 50 variables). Comparing it to the eigenvalue criterion results generated by the test data set seems to validate the results generated by the train data set as discussed in Section 3.3. Eigenvalue criterion on test data set suggested 34 dimensions with an estimated shared percent variance of 83%, which validates the eigenvalue criterion results generated by the train data set (the suggested result was 38 dimensions with an estimated shared percent variance of 80%). The results did not show any huge difference in the dimensions' shared percent variance, and almost the same number of dimensions were selected. It shows that these details validate the results generated by the train data set and now can be reported as Valid.

## 4   Conclusion

Obesity is one of the non-communicable diseases that is a condition of being overweight or a major nutritional disorder. The prevalence of obesity in Brunei Darussalam has increased more than double since 1997, to 27.2% in 2011, and around 61% of Bruneians are overweight and obese, which is highest in the ASEAN region. Comparatively, in the U.S, the prevalence of obesity in 2011-2014 was 22.8% (including obese and extremely obese individuals) among the youth aged 2-19 years which shows that obesity has become a worldwide epidemic. Its growth has been projected at 40% in the upcoming decade. In this study, the classification technique was used to identify the obesity subgroups within the NHANSS data provided by the ministry of health, Brunei Darussalam.

The novelty of the research was to extract useful knowledge from NHANSS data of mixed variable types as not many studies have been carried out in the past in this domain with mixed data types. CATPCA algorithm was used, which grouped the obese sample into two classes concerning the anchoring conditions related to obesity. The two subgroups presented the most prevalent conditions belonging to demographic, socio-economic, smoking, anthropometric, short food frequency characteristics of the obese sample. The short food frequency revealed that the obese group was not taking care of their diet and was used to eating nasi katok (local rice cooked with fried chicken) and chicken Tail / Wings / Skin twice a week. Noticeably the history of blood pressure, diabetes mellitus, and high blood cholesterol were undetected for obese patients, but most of them were reported as having anemia as far as their health was concerned. All of these results were validated, and profiling was noted accordingly.

This research is of clinical importance, and the salient features should be reported and further investigated from a medical perspective. The proposed approach reveals the sub-groups that may help investigate the importance of the lifestyle factors (i.e., age, smoking habits, blood pressure, diabetes mellitus, high blood cholesterol, etc.) from a clinical point of view. Overall, the combination of clinical knowledge with data-hidden information and the evaluation of subclasses revealed by the data structure could lead to very interesting developments.

**Acknowledgments**

The authors would like to express sincere appreciation for the technical assistance and support from the Department of Economic Planning and Development Brunei Darussalam, research assistant, and managers from the Ministry of Health Brunei Darussalam and participation from the survey respondents.

**Conflict of Interest**

The author(s) declared no potential conflicts of interest concerning this article's research, authorship, and/or publication.

**References**

[1]   Brunei, "Brunei Darussalam Government Gazette Part iii

Smoking in Specified Places and Specified Vehicles," Bandar Seri Begawan, Brunei Darussalam, 2005.

[2] L. Uccioli *et al.*, "Autonomic neuropathy and transcutaneous oxymetry in diabetic lower extremities," Geneva - Switzerland, 1994.

[3] Ministry of Health Brunei Darussalam, "Brunei Darussalam National Multisectoral Action Plan for the Prevention and Control of Noncommunicable Diseases 2013-2018," Bandar Seri Begawan, Brunei Darussalam, 2013.

[4] I. A. WM Nazlee WZ, Rosnani Z, "Brunei International," *Brunei Int. Med. J.*, vol. 15, no. April, pp. 53–57, 2019.

[5] I. ASEAN Secretariat, Jakarta, *The ASEAN Secretariat Jakarta*. Jakarta, Indonesia: ASEAN Secretariat,December 2018, 2018.

[6] S. K. Ong *et al.*, "Cross-sectional STEPwise Approach to Surveillance (STEPS) Population Survey of Noncommunicable Diseases (NCDs) and Risk Factors in Brunei Darussalam 2016," *Asia-Pacific J. Public Heal.*, vol. 29, no. 8, pp. 635–648, 2017.

[7] U. Khalil, O. A. Malik, D. Lai, and O. S. King, "Identifying sub-groups of the obese from national health and nutritional status survey data using machine learning techniques," in *IET Conference Publications*, 2018, vol. 2018, no. CP750, pp. 113 (4 pp.)-113 (4 pp.).

[8] C. L. Ogden, M. D. Carroll, B. K. Kit, and M. Flegal, "Ogden, C. L., Carroll, M. D., Kit, B. K., & Flegal, M. (2016). Prevalence of Childhood and Adult Obesity in the United States, 2011–2012," *Jama*, vol. 311, no. 8, pp. 806–814, 2016.

[9] ASEAN Secretariat, *Association of Southeast Asian Nations, Annual Report, 2013-2014*. Jakarta, Indonesia: JAKARATA, ASEAN Secretariat, 2014.

[10] B. MoH, "The Report, The 2nd National Health and Nutritional Status Survey (NHNANSS) 2014.," Ministry of Health, Common Wealth Drive, Brunei Darussalam, Bandar Seri Begawan, 2014.

[11] A. Othman, "Brunei records highest child obesity rate in region | Borneo Bulletin Online," *Borneo Bulletin*, 2020. [Online]. Available: https://borneobulletin.com.bn/brunei-records-highest-child-obesity-rate-in-region-2/. [Accessed: 21-Apr-2019].

[12] D. S. S. D. H. Z. B. H. Hanafi, "Message by Yang Berhormat Dato Seri Setia Dr Haji Zulkarnain Bin Haji Hanafi Minister Of Health On The Occasion Of World Cancer Day 2017," *Moh.Gov.Bn*, 2017. [Online]. Available: http://www.moh.gov.bn/Lists/CO_Announcements/NewDispForm.aspx?ID=52. [Accessed: 21-Apr-2019].

[13] U. Khalil, O. A. Malik, D. Teck, C. Lai, and O. S. King, "Cluster Aanalysis for Identifying Obesity Subgroups in Health and Nutritional Status Survey Data," *Asia-Pacific J. Inf. Technol. Multimed.*, vol. 10, no. 2, pp. 146–169, 2021.

[14] N. Antonioli *et al.*, "Ontology-based data management for the Italian public debt," *Frontiers in Artificial Intelligence and Applications*, 2014. [Online]. Available: http://www.who.int/mediacentre/fac

tsheets/fs311/en/. [Accessed: 20-Feb-2018].

[15] M. A. Green, M. Strong, F. Razak, S. V. Subramanian, C. Relton, and P. Bissell, "Who are the obese? A cluster analysis exploring subgroups of the obese," *J. Public Heal. (United Kingdom)*, vol. 38, no. 2, pp. 258–264, 2016.

[16] A. Ghatak, *Machine Learning with R*, Second edi. Livery Place 35 Livery Street Birmingham B3 2PB, UK., 2017.

[17] I. Kavakiotis, O. Tsave, A. Salifoglou, N. Maglaveras, I. Vlahavas, and I. Chouvarda, "Machine Learning and Data Mining Methods in Diabetes Research," *Comput. Struct. Biotechnol. J.*, vol. 15, pp. 104–116, 2017.

[18] Y. Mori, M. Kuroda, and N. Makino, "Nonlinear Principal Component Analysis and Its Applications SPRINGER BRIEFS IN STATISTICS," *SPRINGER BRIEFS Stat.*, vol. 1, no. June 2016, pp. 2–85, 2016.

[19] J. won Lee and C. Giraud-Carrier, "Results on mining NHANES data: A case study in evidence-based medicine," *Comput. Biol. Med.*, vol. 43, no. 5, pp. 493–503, 2013.

[20] C. A. Befort, N. Nazir, and M. G. Perri, "Behavior Risk Factor Surveillance System (BRFSS) 5 and the 1997-1998 National Health J Rural Health," *J Rural Heal.*, vol. 28, no. 4, pp. 392–397, 2012.

[21] A. J. (2007). Linting, M., Meulman, J.J., Groenen, P.J.F., & Van der Kooij, "Nonlinear Principal Components Analysis :," *Am. Psychol. Assoc.*, no. 2007, pp. 12–48, 2004.

[22] M. Linting, J. J. Meulman, P. J. F. Groenen, and A. J. van der Kooij, "Nonlinear Principal Components Analysis: Introduction and Application," *Psychol. Methods*, vol. 12, no. 3, pp. 336–358, 2007.